\documentclass[10pt,journal,compsoc]{IEEEtran}
\ifCLASSOPTIONcompsoc
  \usepackage[nocompress]{cite}
\else
  \usepackage{cite}
\fi

\usepackage{amsmath,amssymb,amsfonts}
\usepackage[pdftex]{graphicx}
\usepackage{url}
\usepackage{subfigure,stfloats}
\usepackage{booktabs}
\usepackage{hyperref}
\hypersetup{
    colorlinks=true,
    linkcolor=blue,
    filecolor=magenta,      
    urlcolor=cyan,
}

\graphicspath{{images/}}

\begin{document}
\bstctlcite{IEEEexample:BSTcontrol}
\title{On the use of uncertainty in classifying \textit{Aedes Albopictus} mosquitoes}

\author{Gereziher~Adhane,
        Mohammad Mahdi~Dehshibi,
        and~David~Masip,~\IEEEmembership{Senior~Member,~IEEE}
\IEEEcompsocitemizethanks{\IEEEcompsocthanksitem Department of Computer Science, Universitat Oberta de Catalunya, 08018, Barcelona, Spain.\protect\\
Corresponding author: Gereziher Adhane (e-mail: gadhane@uoc.edu).}
\thanks{Manuscript is published.~\url{https://doi.org/10.1109/JSTSP.2021.3122886}}}

\markboth{IEEE Journal of Selected Topics in Signal Processing}%
{Adhane \MakeLowercase{\textit{et al.}}: On the use of uncertainty in classifying \textit{Aedes Albopictus} mosquitoes}

\IEEEtitleabstractindextext{%
\begin{abstract}
The re-emergence of mosquito-borne diseases (MBDs), which kill hundreds of thousands of people each year, has been attributed to increased human population, migration, and environmental changes. Convolutional neural networks (CNNs) have been used by several studies to recognise mosquitoes in images provided by projects such as Mosquito Alert to assist entomologists in identifying, monitoring, and managing MBD. Nonetheless, utilising CNNs to automatically label input samples could involve incorrect predictions, which may mislead future epidemiological studies. Furthermore, CNNs require large numbers of manually annotated data. In order to address the mentioned issues, this paper proposes using the Monte Carlo Dropout method to estimate the uncertainty scores in order to rank the classified samples to reduce the need for human supervision in recognising \textit{Aedes albopictus} mosquitoes. The estimated uncertainty was also used in an active learning framework, where just a portion of the data from large training sets was manually labelled. The experimental results show that the proposed classification method with rejection outperforms the competing methods by improving overall performance and reducing entomologist annotation workload. We also provide explainable visualisations of the different regions that contribute to a set of samples' uncertainty assessment.
\end{abstract}

}

\maketitle
\IEEEraisesectionheading{\section{Introduction}\label{sec:introduction}}
\IEEEPARstart{R}{ecent} studies indicate that insects play a relevant role in propagating and outbreaking infectious and contagious diseases~\cite{35ogawa2015seroepidemiological,36verhagen2015virus}. Just a few dozen of the over 3600 recognised mosquito species, including \textit{Aedes aegypti} and \textit{Aedes albopictus}, can carry and transmit mosquito-borne diseases (MBDs) to humans through direct or indirect contact~\cite{wilkerson2015making,franklinos2019effect}. These mosquitoes are often referred to as the world's deadliest species~\cite{jit2016economic} since they cause hundreds of thousands of deaths each year by transmitting diseases such as yellow fever, dengue fever, Zika, and chikungunya~\cite{bhatt2013global,petersen2016chikungunya,weaver2015chikungunya}.

Due to the lack of effective vaccines for MBDs~\cite{manning2018mosquito}, urgent improvement of global disease vector monitoring and control (both invasive and native) across colonised territories is considered as a strategic approach to disease prevention and outbreak response~\cite{roiz2018integrated,bartumeus2019sustainable}. Mosquito Alert\footnote{www.mosquitoalert.com/en}~\cite{27} is a platform launched by a group of Spanish academic institutions with the goal of raising public awareness and establishing expert-validated citizen science networks to monitor and control MBDs. This platform brings together citizens, entomologists, public health professionals, and mosquito control services to assist minimise MBDs in Spain. The data collection process in this platform relies on people uploading geo-tagged photos of mosquitoes and mosquito breeding sites taken in the field using a dedicated smartphone app. Images are added to the database and shared with control services and public health departments after inspecting and validating by a team of entomologists.

Although this platform simplifies the traditional and labour-intensive method of mosquito surveillance (collecting authoritative data by trapping adults, dipping for larvae), the system receives an increase in the number of uploaded images. The received reports may also contain one or more images that should be manually inspected by expert entomologists to select the best picture. In the long run, manual inspection of each image may not be easily scalable because providing human annotations is time-consuming and costly. As a result, this methodology could be improved by utilising automated algorithms capable of making reliable predictions for a well-defined dataset.
Several studies have used state-of-the-art deep learning or other forms of machine learning approaches~\cite{adhane2021deep,pataki2021deep,fernandes2021detecting,park2020classification,9462559,38okayasu2019vision,fuad2018training,39lorenz2015artificial} to minimise the workload of human annotators, make the system scalable and flexible, and increase the fitness of use~\cite{palmer2017citizen}.

This paper proposes two significant contributions to minimise human intervention while maximising model accuracy in the citizen science project with a large amount of contributed data. First, we included the Monte Carlo (MC) dropout~\cite{2gal2016dropout} to our fine-tuned version of VGG16~\cite{adhane2021deep} to incorporate uncertainty in the prediction. To do so, a dropout layer with a probability of $\alpha=0.5$ was added after each convolution layer, and the dropout rate of the fully-connected layer was set to $\beta=0.4$. During prediction, we keep dropout active and make $T$ stochastic forward passes through the network. Averaging the softmax vector obtained from $T$ forward passes provides the final posterior probability distribution for a given input, where the variance of this vector serving as model uncertainty. In this way, samples that may have been misclassified can be identified. The predicted samples were then ranked, and only samples with the highest level of uncertainty were referred to an expert. Second, we propose an active learning scheme that initiates with a small set of manually annotated samples and gradually adds more unlabelled data. The uncertainty estimation guides this process and sends only the labels that are more likely to be inaccurate to the expert. 

Incorporating the uncertainty measure to the CNN output had a threefold advantage in the automated classification of mosquito images: (1) it reduced the number of misclassified samples by sorting the most uncertain samples at the time of inference and requesting re-validation by entomologists ; (2) it reduced the need for annotation of the entire database by assisting annotators in an active learning framework during training; and (3) it improved the overall performance of our fine-tuned version of VGG16~\cite{adhane2021deep} architecture by 4\%. Furthermore, we analysed the causes of model uncertainty by highlighting regions in certain and uncertain predictions using a Bayesian visual explanation method~\cite{45bykov2020much}. The model highlighted the abdomen, thorax, and legs of the \textit{Aedes albopictus} in the most certain predictions, while the non-discriminant regions in the predictions associated with high variance were highlighted.

\begin{figure}[!htbp]
    \xdef\xfigwd{\columnwidth}
    \centering
    \includegraphics[width=\linewidth]{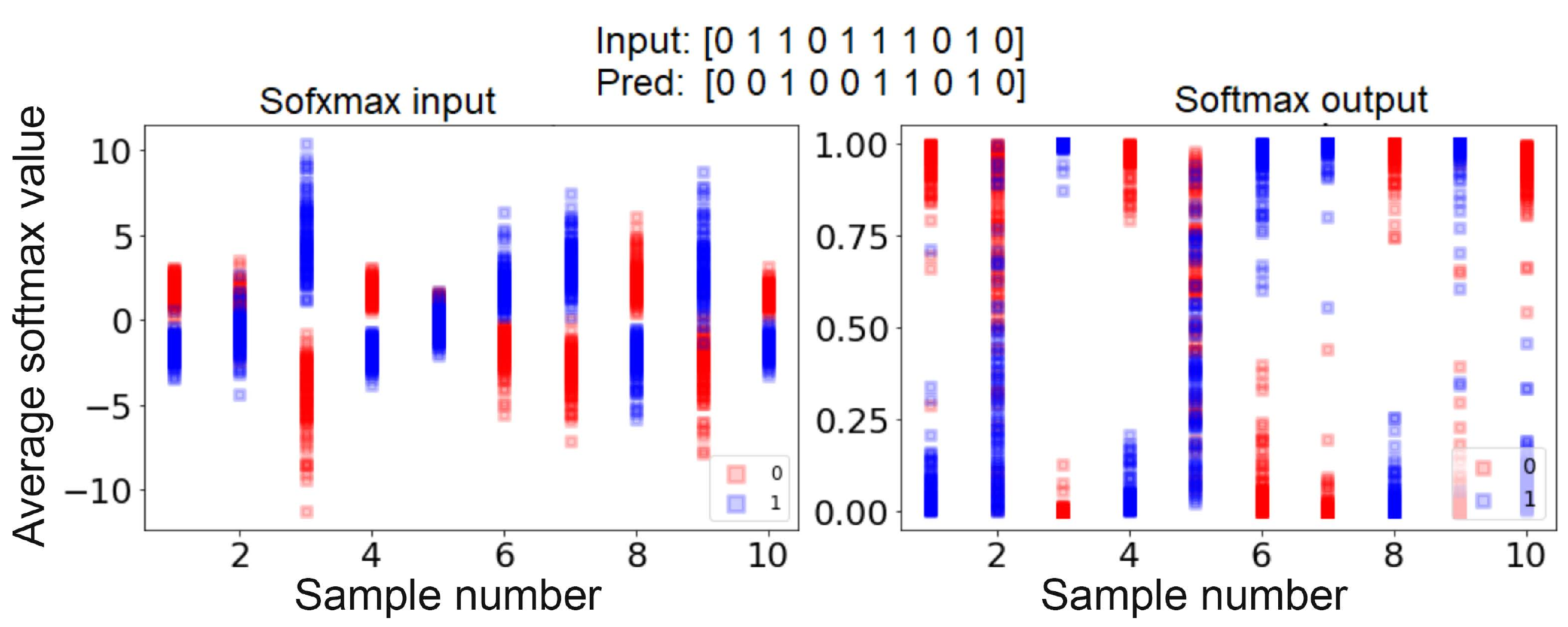}
    \caption{A scatter of $T = 100$ forward passes of softmax input and the final posterior predictive distribution for ten samples. In this example, input 2 and 5 have a high level of model uncertainty, resulting in an incorrect prediction.}
    \label{fig:00}
\end{figure}

To exemplify the proposed approach, we show a scenario with ten inputs belonging to two classes, where blue represents class 1 and red represents class 0, as shown in Fig.~\ref{fig:00}. We evaluated the network trained on the Mosquito Alert data set by performing $T=100$ stochastic forward passes through the trained network. In this scenario, when the softmax inputs were well-differentiated, the output class can be confidently predicted. However, where the distribution of the input features for two classes overlaps (e.g., inputs 2 and 5) the estimated uncertainty is high, resulting in an inaccurate prediction. In this way, we can submit the most uncertain samples to experts for further evaluation and improve the overall prediction accuracy, accordingly.

The rest of this paper is organised as follows: Section~\ref{sec:relatedwork} surveys the previous studies. The proposed methodology is detailed in Section~\ref{sec:methodology}. Section~\ref{sec:experiment} presents experimental results. Finally, Section~\ref{sec:conclusion} concludes the paper.


\section{Related Work}\label{sec:relatedwork}
In this section, we review previous research from two perspectives: (i) mosquito classification using CNN, and (ii) leveraging uncertainty in CNN prediction.

\noindent\textbf{-- Mosquito classification}: In~\cite{61favret2016machine,62fuchida2017vision}, researchers proposed to use a support vector machine to identify mosquito images from other insects such as bees and flies. Minakshi et al.~\cite{63minakshi2018leveraging} classified images of nine different mosquito species collected in-the-wild using a multi-class support vector machine. This study was then extended by developing a mask R-CNN model to extract and detect the key anatomical components of different mosquito species, such as the thorax, wings, abdomen, and legs~\cite{60minakshi2020framework}.

Schreiber et al.~\cite{fernandes2021detecting} used CNN to identify adult \textit{Aedes aegypti} mosquito species using wingbeat audio recorded by smartphones in a nearly silent environment. They trained binary, multiclass, and ensemble of binary classifiers using the recorded spectrogram to represent the frequency of mosquito wing-beats over time. However, because the wingbeat recordings were made in a low-noise environment, the performance of classifiers in processing noisy recordings may be reduced when employed in large-scale environment. Furthermore, this technique is costly due to the high cost of optical sensors and the difficulty in obtaining high-quality audio recordings.

Researchers suggested CNN-based mosquito classification techniques for deciding if a mosquito larva belongs to the vector species in~\cite{fuad2018training,sanchez2017mosquito}. Fine-tuning pre-trained CNNs with a small dataset resulted in high classification accuracy. However, since the data collection and experiments were carried out in a laboratory setting, taxonomists and health workers cannot employ these methods in the field studies. Park et al.~\cite{park2020classification} used transfer learning to fine-tune a pre-trained CNN to classify six distinct mosquito species with similar morphological structures. While they reported high classification accuracy, which could also locate the discriminative regions of mosquitoes, the use of the data augmentation technique to increase the number of samples decrease the model's generalisability for analysing images collected in-the-wild.

Adhane et al.~\cite{adhane2021deep} proposed to use transfer learning to fine-tune Deep Convolutional Neural Networks on Mosquito Alert data set~\cite{27} to automate the classification of \textit{Aedes albopictus} mosquito images. They also used Grad-CAM visualisation to highlight the most discriminant regions of \textit{Aedes albopictus} images, which coincided with the white band stripes found on the legs, abdomen, and thorax. Grad-CAM visualisation of \textit{Aedes albopictus} discriminant regions aided in analysing subsequent classification errors, which were linked to poor acquisition conditions and large image occlusions. Furthermore, Pataki et al.~\cite{pataki2021deep} developed a deep learning model using the Mosquito Alert data set~\cite{27} to cover certain aspects of the data such image quality, quantity, geographic diversity, and usefulness. 

Although increasing the number of annotated data and GPU performance can improve CNN results for MBD classification task~\cite{65kittichai2021deep,66kim2019deep}, the lack of prediction reliability can limit their use.  To address this issue, leveraging uncertainty into the prediction probabilities for ambiguous case analysis will significantly improve the performance of such methods.

\noindent\textbf{-- Uncertainty in prediction}: Several studies have explored uncertainty estimation in various deep learning models for various image processing applications such as object detection~\cite{46harakeh2020bayesod}, segmentation~\cite{49martinez2019segmentation}, human pose detection~\cite{50gundavarapu2019structured}, facial recognition~\cite{53zheng2019uncertainty}.

Stoean et al.~\cite{54stoean2020ranking} analysed electrooculography time series data using Monte Carlo Dropout for uncertainty quantification in deep neural network (DNN) models. They used a convolutional long short-term memory network with a Monte Carlo dropout layer to measure the learning model's uncertainty. They estimated the mean and standard deviation for each Monte Carlo estimate, serving as the confidence to perform both classification and ranking. They validated their approach on real-world data and reported an improved average classification accuracy.

By incorporating a learnable parameter, Khairnar et al.~\cite{55khairnar2020modified} proposed a Bayesian CNN with an adaptive activation function to dynamically adjust the loss function, increasing the accuracy and convergence rate in breast histopathology image classification. Dropout-based Bayesian uncertainty measure was evaluated by Leibig et al.~\cite{31leibig2017leveraging} for estimating diabetic retinopathy from fundus images. They showed that DNNs' measured uncertainty is useful for ranking predictions.

Li et al.~\cite{li2020uncertainty} investigated uncertainty calibration within active learning framework for medical image segmentation and found more effective with minimum labelling effort. Several uncertainty assessment methods and acquisition techniques for active learning approaches have also been investigated on tasks such as image classification~\cite{gal2017deep} and visual question answering~\cite{patro2019u}, which demonstrated significant improvement with minimum labelling.

\section{Proposed Method}\label{sec:methodology}
Quantifying uncertainty in CNNs, particularly during the decision-making process, can not only improve classification results but also reduce labelling effort for experts when the trained model is used in scenarios such as classification with rejection and active learning. In this work, inspired by~\cite{2gal2016dropout}, we used the Monte Carlo (MC) dropout to quantify model uncertainty for the classification of \textit{Aedes albopictus} images. We also integrated uncertainty guided continuous learning into active learning acquisition functions to explore uncertain samples in the prediction.

\subsection{Uncertainty in classification with rejection}
Given a dataset of $n$ samples $\mathcal{D} = \{(x_{i}, y_{i})\}_{i=1}^{n}$, with $x_{i} \in \mathbb{R}^{d_{x}}$ and $y_{i} \in \mathbb{R}^{d_{y}}$, our goal is to train a neural network $\mathcal{H}(x) = \mathbb{E}[\mathbf{Y}|\mathbf{X} = x]$ to classify \textit{Aedes albopictus} mosquito images by minimising the cross-entropy between the class labels and the softmax output as in Eq.~\ref{eq:01}

\begin{equation}
    \label{eq:01}
    p(y_i|x; w,b) = \frac{\exp(x^T w_i+b_i) }{\sum_{j\in d_{y}}\exp(x^T w_j+b_j)}
\end{equation}

In this study, Mosquito Alert dataset~\cite{27} was used in which the input image was labelled by $d_{y} = \{0,1\}$. The input image $\mathbf{X}$ was scaled to the size of $d_{x} = 224 \times 224 \times 3$. The output tag $\tilde{y}_{i} = 0$ if the $i$-th sample was \textit{Aedes albopictus}, and $\tilde{y}_{i} =1 $ otherwise.

The class probability obtained from this softmax output is a point estimate, and it is often interpreted as model confidence~\cite{guo2017calibration}. However, the point estimate approach does not provide an uncertainty measure on the predictions and tends to be overconfident for out-of-distribution samples~\cite{nixon2019measuring}, particularly in our case where mosquito bodies have roughly similar micro patterns. Bayesian Neural Network can provide a better understanding of the uncertainty associated with underlying processes.

We integrated the concept of Monte Carlo (MC) Dropout into the fine-tuned version of VGG16~\cite{adhane2021deep} to obtain a calibrated model with uncertainty estimation, as motivated by~\cite{gal2017deep,li2017dropout}. After each convolution and fully-connected layer, we added a dropout layer with a probability of $\alpha$ and $\beta$, respectively, and kept these layers active during the evaluation phase to define a variational posterior distribution for each weight matrix, as shown in Eq.~\ref{eq:05}.

\begin{equation}
    \label{eq:05}
    \begin{aligned}
        z_{i} & \sim Bernoulli(p_{i})\\
        W_{i}& = M_{i} \cdot diag(z_{i}),
    \end{aligned}
\end{equation}
where $z_{i}$ denotes the random inactivation coefficients, $M_{i}$ denotes the weights matrix prior to dropout, and $p_{i}$ denotes the activation probability for the $i^{th}$ layer, which can be learned or manually set. Equation~\ref{eq:06} expresses the equivalence between a standard objective function used for training with dropout and additional weight regularisation, where $\lambda$ is the regularisation parameter.
\begin{align}
    \label{eq:06}
        \ell_{droupout} = - \sum_{i=1}^{n}\log  \frac{\exp(x_{i}^{T}w_i+b_i)}{\sum_{j \in d_{y}}\exp(x_{i}^{T}w_j+b_j)}\\ \nonumber
        + \lambda \sum_{i=1}^{n}w_{i}^{2}.
\end{align}

We performed $T$ stochastic forward passes through the trained network during evaluation to implement the MC Dropout. Each stochastic forward pass ($t \in \{1,2,\cdots,T\}$) produces a new softmax prediction ($\tilde{y}^{t}$). Averaging softmax outputs provides the final posterior predictive distribution for the input $x$, where the distribution's variance ($\sigma$) was used to represent model uncertainty, see Eq.~\ref{eq:07}.

\begin{equation}
    \label{eq:07}
    \begin{aligned}
        \mu =\frac{1}{T}\sum_{i=1}^{T}\tilde{y}^{i},~~ \sigma = \frac{1}{T-1}\sqrt{\sum_{i=1}^{T} \left ( \tilde{y}^{i} -\mu \right )^{2}}.
    \end{aligned}
\end{equation}

We used $\sigma$ to exclude predictions with a high degree of uncertainty, i.e., classification with the probability of rejection, and formulate the uncertainty-based classification with rejection as expressed in Eq.~\ref{eq:08}.

\begin{equation}
    \label{eq:08}
    \tilde{y} = \left\{\begin{matrix}
    \mathcal{H}(x) & \sigma \leq \tau\\ 
    \text{reject}  & \sigma > \tau
    \end{matrix}\right.
\end{equation}
where $\tau$ is the rejection point's threshold value, and we discuss its impact in Section~\ref{sec:experiment}.

\subsection{Uncertainty in active learning framework}
Active learning is suitable for problems where unlabelled data is readily available, but sample annotation is expensive. It enables the classifier to learn from a smaller number of the most informative samples while still achieving satisfactory results. In the proposed classification with rejection, the estimated uncertainty was used as an acquisition strategy in the active learning framework. The variance of the softmax vector $\tilde{y}^{t}$ obtained from $t \in \{1, 2, \cdots, T\}$ forward passes was used in this framework to rank the dataset pool and refer samples with the highest variance (i.e., most uncertain prediction) to an expert.

More formally, we first trained the model ($\mathcal{H}$) with samples belong to the initial labelled dataset ($\mathcal{D}_{l}$) and validated it on the validation set ($\mathcal{D}_{v}$). The model then used Eq.~\ref{eq:07} to estimate the $\sigma$ for all samples belonging to unlabelled pool of data ($\mathcal{D}_{u}$) and selected $\kappa << \parallel \mathcal{D}_{u} \parallel$ samples that satisfy $\sigma > \tau$, i.e., samples of which the model is uncertain. We asked the expert to provide labels for these $\kappa$ samples, which we then added to $\mathcal{D}_{l}$. The model was again fine-tuned with the $\kappa$ updated labelled data and validated on $\mathcal{D}_{v}$. To avoid overfitting, we used the early stopping strategy with a validation set during the training of the active learning framework. Therefore, we terminate training if the model's performance is close to or greater than the project goal. Otherwise, the procedure is repeated until the expert has labelled all of the data $\mathcal{D}_{u}$.

\section{Experiments}\label{sec:experiment}
\subsection{Mosquito Alert dataset}
We conducted experiments on the open-source dataset from the Mosquito Alert project\footnote{The curated database is available on http://www.mosquitoalert.com/en/mosquito-images-dataset}~\cite{27}. This project was launched in 2014 by the Center for Research and Ecological Applications (CREAF) and the Center for Advanced Studies of Blanes (CEAB-CSIC) near Barcelona (Spain) to monitor and control disease-carrying mosquitoes. The data collection process for this platform is based on images of mosquitoes and mosquito breeding sites submitted by volunteers. The uploaded images are in RGB format. Since this platform has no photo size restrictions, image sizes range from $200\times200\times3$ to $460\times460\times3$, with an average size of $420\times368\times3$. A team of three entomologists inspects, validates, and classifies the images that are submitted.  In the event of a disagreement, the final label is assigned by a super-expert who holds the final decision ability.

\begin{figure}[!htbp]
    \centering
    \includegraphics[width=\linewidth]{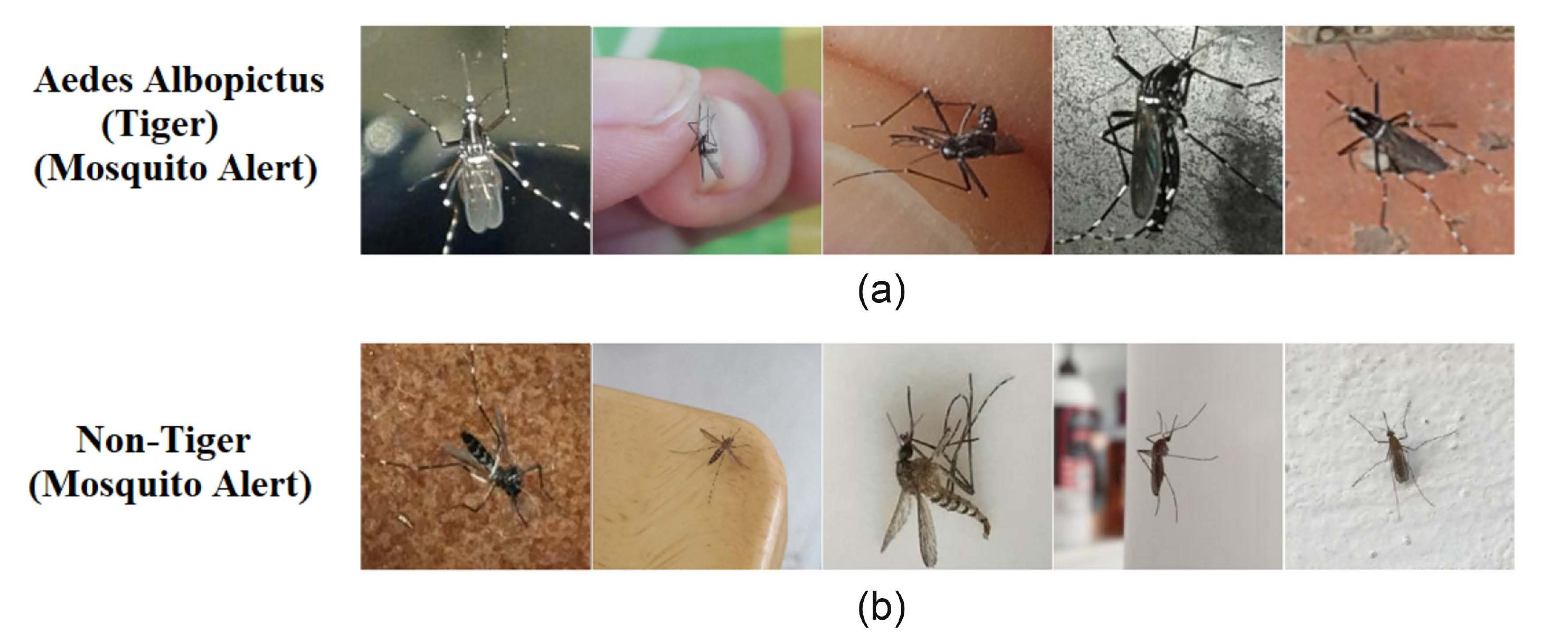}
    \caption{Sample of (a) \textit{Aedes albopictus} (tiger) and (b) \textit{non-tiger} mosquitoes from the Mosquito Alert data set.}
    \label{fig:samples}
\end{figure}

Each image uploaded is labelled as \textit{Aedes albopictus}, \textit{Aedes aegypti}, other species, unclassified or unknown. The images of mosquito breeding sites do not yet have tags. Each image, in each category, is classified as confirmed, or probable by the experts. The Mosquito Alert data set contains a total of 3364 confirmed \textit{Aedes albopictus} images. Fig.~\ref{fig:samples} shows ten samples from the Mosquito Alert dataset, five of which are \textit{Aedes albopictus} and the other five are \textit{Non-tiger} samples.

In this study, we used 3364 images labelled as confirmed \textit{Aedes albopictus} cases as positive samples (i.e., tiger), and other species as negative samples, i.e., non-tiger. As a result, the architecture was trained using a total number of 6378 tiger and non-tiger images.

\subsection{Architecture details}
\begin{figure}[!htbp]
    \xdef\xfigwd{\columnwidth}
    \centering
    \includegraphics[width=0.9\linewidth]{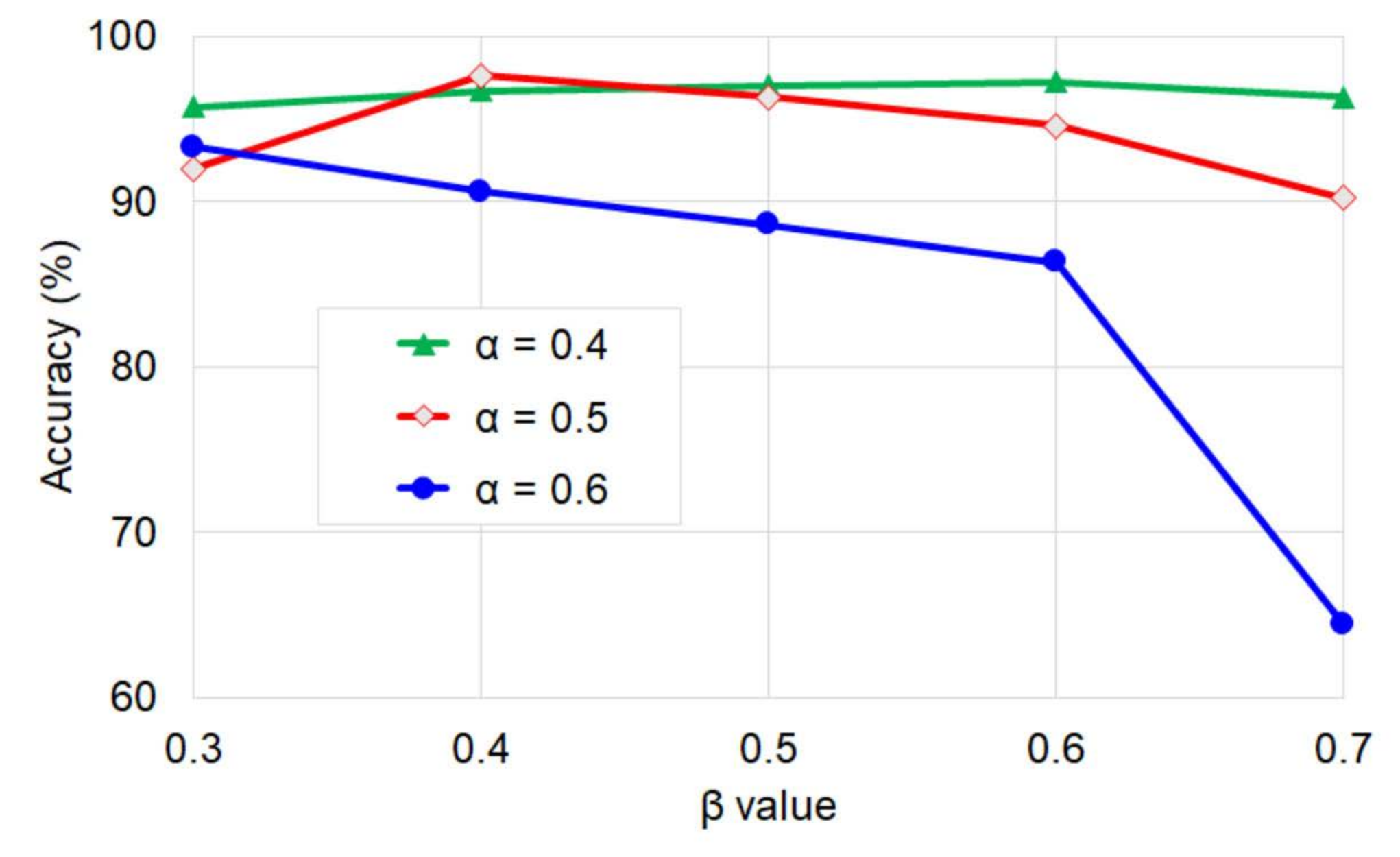}
    \caption{Accuracy of the proposed architecture as a function of $\alpha$ and $\beta$, where $\alpha$ and $\beta$ are the dropout rates for the hidden and fully-connected layers, respectively.}
    \label{fig:09}
\end{figure}

We modified our previous contribution to the VGG16~\cite{adhane2021deep} by adding a dropout layer with the probability of $\alpha$ after each convolution layer and setting the dropout rate of the fully connected layers to $\beta$ to measure uncertainty in the underlying prediction. We trained the proposed architecture as a function of $(\alpha, \beta)$ using five-fold cross-validation and plotted the average accuracy to find the best combination of these two dropout rates (see Fig.~\ref{fig:09}). An accuracy value of 97.6\% at $\alpha = 0.5$ and $\beta =0.4$ was obtained, and these values were used in the rest of the experiments. The proposed architecture was trained end-to-end in PyTorch~\cite{paszke2019pytorch} using the Tesla K80 GPU\footnote{The implementations are available at  https://github.com/ageryw/UncertaintyInMosquitoClassification}.

We used cross-entropy as a loss function and stochastic gradient descent with initial learning rates and momentum of 0.001 and 0.9, respectively, during the training process. The learning rate decayed by a factor of 0.1 every 5 epochs by using the StepLR scheduler. We fed the network with a mini-batch size of 64, and the optimisation process was terminated after 25 epochs. We did not use data augmentation to avoid unrealistic changes in micro-morphological patterns of mosquitoes that could have skewed the final results. We kept the dropout active during the test phase to measure the uncertainty outcomes from the MC dropout by conducting $T=100$ stochastic forward passes through the network~\cite{gal2015bayesian}. As a result, rather than a single point estimate for a given input, we have a per-class output distribution of the softmax confidence in which the distribution variance serves as the model uncertainty.

\begin{table*}[!htbp]
\centering
\caption{The performance of informed referral of uncertain samples for different architectures are compared using class-wise recall, precision, and F1 score. Here, $\sigma$ represents the standard deviation and TP, TN, FP and FN stand for True Positive, True Negative, False Positive and False Negative, respectively.}
\label{tab:01}
\begin{tabular}{@{}lcccccccccc@{}}
\toprule
\textbf{Model}                 & \multicolumn{1}{l}{\textbf{Un. Threshold}} & \multicolumn{1}{l}{\textbf{\begin{tabular}[c]{@{}l@{}}Rejected\\  samples \#\end{tabular}}} & \multicolumn{1}{l}{\textbf{\begin{tabular}[c]{@{}c@{}}Retained samples\\ (Samples \#)\end{tabular}}} & \multicolumn{1}{l}{\textbf{TP}} & \multicolumn{1}{l}{\textbf{TN}} & \multicolumn{1}{l}{\textbf{FP}} & \multicolumn{1}{l}{\textbf{FN}} & \multicolumn{1}{l}{\textbf{Precision}} & \multicolumn{1}{l}{\textbf{Recall}} & \multicolumn{1}{l}{\textbf{F1 score}} \\ \midrule
\textbf{Proposed architecture} & $\sigma \leq$ 0.08                                        & 467                                                                                          & 808                                                                                                      & 502                             & 290                             & 3                               & 13                              & 0.99                                   & 0.98                                & 0.98                                  \\
                               & $\sigma \leq$ 0.1                                         & 390                                                                                          & 885                                                                                                      & 535                             & 329                             & 6                               & 15                              & 0.99                                   & 0.97                                & 0.98                                  \\
\textbf{}                      & $\sigma \leq$ 0.2                                         & 125                                                                                          & 1140                                                                                                     & 636                             & 447                             & 15                              & 52                              & 0.97                                   & 0.93                                & 0.95                                  \\
                               & $\sigma \leq$ 0.3                                         & 3                                                                                            & 1272                                                                                                     & 665                             & 495                             & 30                              & 82                              & 0.96                                   & 0.89                                & 0.92                                  \\
                               & No $\sigma$                                       & 0                                                                                            & 1275                                                                                                     & 664                             & 492                             & 31                              & 88                              & 0.95                                   & 0.88                                & 0.92                                  \\
                               &                                             &                                                                                              &                                                                                                          &                                 &                                 &                                 &                                 &                                        &                                     &                                       \\
\textbf{Adhane et al.~\cite{adhane2021deep}}                 & $\sigma \leq$ 0.08                                        & 195                                                                                          & 1080                                                                                                     & 572                             & 470                             & 20                              & 18                              & 0.97                                   & 0.97                                & 0.97                                  \\
                               & $\sigma \leq$ 0.1                                         & 155                                                                                          & 1120                                                                                                     & 596                             & 477                             & 24                              & 23                              & 0.96                                   & 0.96                                & 0.96                                  \\
\textbf{}                      & $\sigma \leq$ 0.2                                         & 2                                                                                            & 1273                                                                                                     & 651                             & 535                             & 42                              & 45                              & 0.94                                   & 0.94                                & 0.94                                  \\
                               & $\sigma \leq$ 0.3                                         & 0                                                                                            & 1275                                                                                                     & 653                             & 543                             & 42                              & 46                              & 0.94                                   & 0.93                                & 0.94                                  \\
                               & No $\sigma$                                       & 0                                                                                            & 1275                                                                                                     & 652                             & 533                             & 43                              & 47                              & 0.94                                   & 0.93                                & 0.94                                  \\
                               & \multicolumn{1}{l}{}                        & \multicolumn{1}{l}{}                                                                         & \multicolumn{1}{l}{}                                                                                     & \multicolumn{1}{l}{}            & \multicolumn{1}{l}{}            & \multicolumn{1}{l}{}            & \multicolumn{1}{l}{}            & \multicolumn{1}{l}{}                   & \multicolumn{1}{l}{}                & \multicolumn{1}{l}{}                  \\
\textbf{AlexNet~\cite{krizhevsky2012imagenet}}               & $\sigma \leq$ 0.08                                        & 254                                                                                          & 1021                                                                                                     & 593                             & 377                             & 15                              & 36                              & 0.97                                   & 0.94                                & 0.95                                  \\
                               & $\sigma \leq$ 0.1                                         & 220                                                                                          & 1055                                                                                                     & 606                             & 390                             & 16                              & 43                              & 0.97                                   & 0.93                                & 0.95                                  \\
                               & $\sigma \leq$ 0.2                                         & 37                                                                                           & 1238                                                                                                     & 652                             & 473                             & 34                              & 79                              & 0.95                                   & 0.89                                & 0.92                                  \\
                               & $\sigma \leq$ 0.3                                         & 0                                                                                            & 1275                                                                                                     & 658                             & 487                             & 37                              & 93                              & 0.95                                   & 0.87                                & 0.91                                  \\
                               & No $\sigma$                                       & 0                                                                                            & 1275                                                                                                     & 658                             & 497                             & 37                              & 88                              & 0.95                                   & 0.88                                & 0.91   
                               \\
                               & \multicolumn{1}{l}{}                        &                                                                                              &                                                                                                          &                                 &                                 &                                 &                                 &                                        &                                     &                                       \\
\textbf{Modified AlexNet}    & $\sigma \leq$ 0.08                                        & 705                                                                                          & 570                                                                                                      & 303                             & 251                             & 6                               & 10                              & 0.98                                   & 0.96                                & 0.97                                  \\
                               & $\sigma \leq$ 0.1                                         & 615                                                                                          & 660                                                                                                      & 365                             & 272                             & 9                               & 14                              & 0.97                                   & 0.96                                & 0.97                                  \\
                               & $\sigma \leq$ 0.2                                         & 202                                                                                          & 1073                                                                                                     & 574                             & 402                             & 23                              & 74                              & 0.96                                   & 0.88                                & 0.92                                  \\
                               & $\sigma \leq$ 0.3                                         & 17                                                                                           & 1258                                                                                                     & 636                             & 467                             & 48                              & 107                             & 0.92                                   & 0.85                                & 0.89                                  \\
                               & No $\sigma$                                       & 0                                                                                            & 1275                                                                                                     & 641                             & 471                             & 54                              & 109                             & 0.92                                   & 0.85                                & 0.88                                  \\ \bottomrule
\end{tabular}
\end{table*}

\subsection{Evaluation metrics}
\begin{figure}[!htbp]
    \xdef\xfigwd{\columnwidth}
    \centering
    \subfigure[]{\includegraphics[width=0.31\linewidth]{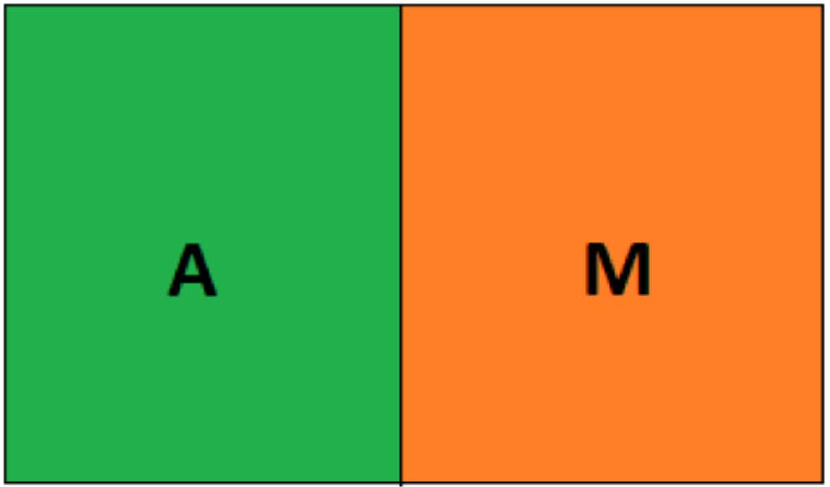}}
    \subfigure[]{\includegraphics[width=0.31\linewidth]{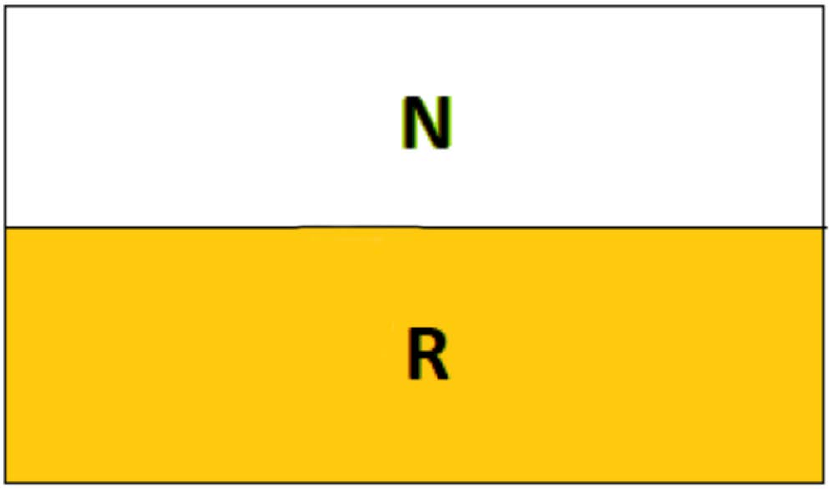}}
    \subfigure[]{\includegraphics[width=0.31\linewidth]{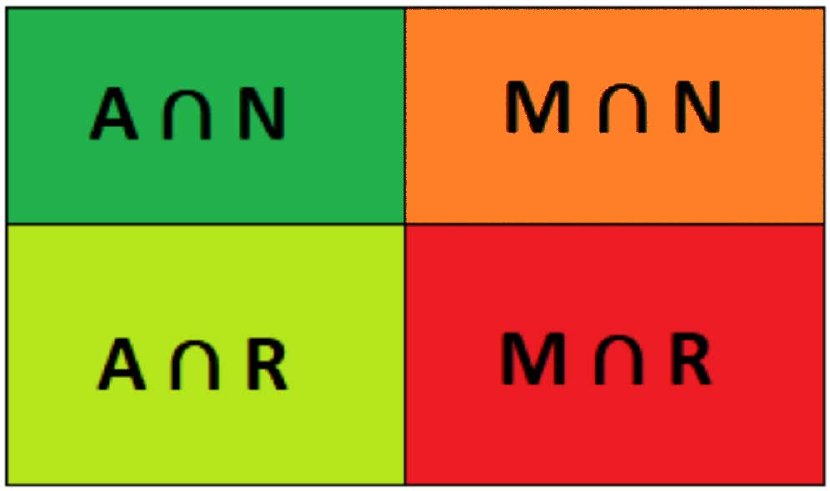}}
    \caption{Rejection performance metrics proposed in~\cite{44condessa2017performance}. (a) Classification partition space (b) rejection partition space, (c) classification with rejection. Correctly classified samples, misclassified samples, non-rejected samples, and rejected samples are represented by the letters $A$, $M$, $N$, and $R$, respectively.}
    \label{fig:03}
\end{figure}

We used the performance metrics proposed by Filipe et al.~\cite{44condessa2017performance} to evaluate the proposed classifier with rejection, which included non-rejection accuracy ($NRA$), classification quality ($CQ$), and rejection quality ($RQ$) to determine the optimal rejection rate, see Fig.~\ref{fig:03}. We formulated the rejection quality metrics in Eq.~\ref{eq:09} using terminologies introduced in~\cite{44condessa2017performance}.

\begin{equation}
    \label{eq:09}
    \begin{aligned}
        NRA &= \frac{\left |A\cap N\right |}{\left | N \right |} \\
        CQ & = \frac{\left |A\cap N\right |+\left |M\cap R\right |}{\left | N \right |+\left | R \right |} \\
        RQ & = \frac{\left |M\cap R\right |\left |A\right |}{\left | A \cap R\right |\left | M \right |}.
    \end{aligned}
\end{equation}
where $A$, $M$, $N$, and $R$ represent correctly classified samples, misclassified samples, non-rejected samples, and rejected samples, respectively. $NRA$ measures the classifier's ability to accurately classify non-rejected samples, $CQ$ measures the classifier's ability to accurately classify non-rejected samples and reject misclassified samples, and $RQ$ measures the classifier's ability to concentrate all misclassified samples into the rejected partition of samples. We also reported precision, recall, and F1-score using Eq.~\ref{eq:10} to provide further insight into the predictive model performance and to compare it to competing methods.

\begin{align}
    \label{eq:10}
    \text{Precision} = \frac{TP}{TP + FP}, ~~ \text{Recall} = \frac{TP}{TP + FN}, \nonumber \\
    \text{F1-score} = 2\times \frac{\text{Precision} \times \text{Recall}}{\text{Precision} + \text{Recall}}.
\end{align}
where $TP$, $FP$ and $FN$ stand for true positive, false positive and false negative, respectively.

\subsection{Results}
We assessed our approach in the following experiments to demonstrate (1) the improvement in classification with rejection as a result of incorporating MC dropout with the probability of $\alpha$ and $\beta$; (2) the comparison of the proposed method with state-of-the-art CNN architectures; (3) the explainability in the classification with rejection; and (4) the use of MC dropout in an active learning framework to minimise annotation effort while maximising performance.

\subsubsection{Impact of uncertainty on classification with rejection}
\begin{figure*}[!htbp]
    \xdef\xfigwd{\columnwidth}
    \centering
    \subfigure[]{\includegraphics[width=0.44\linewidth]{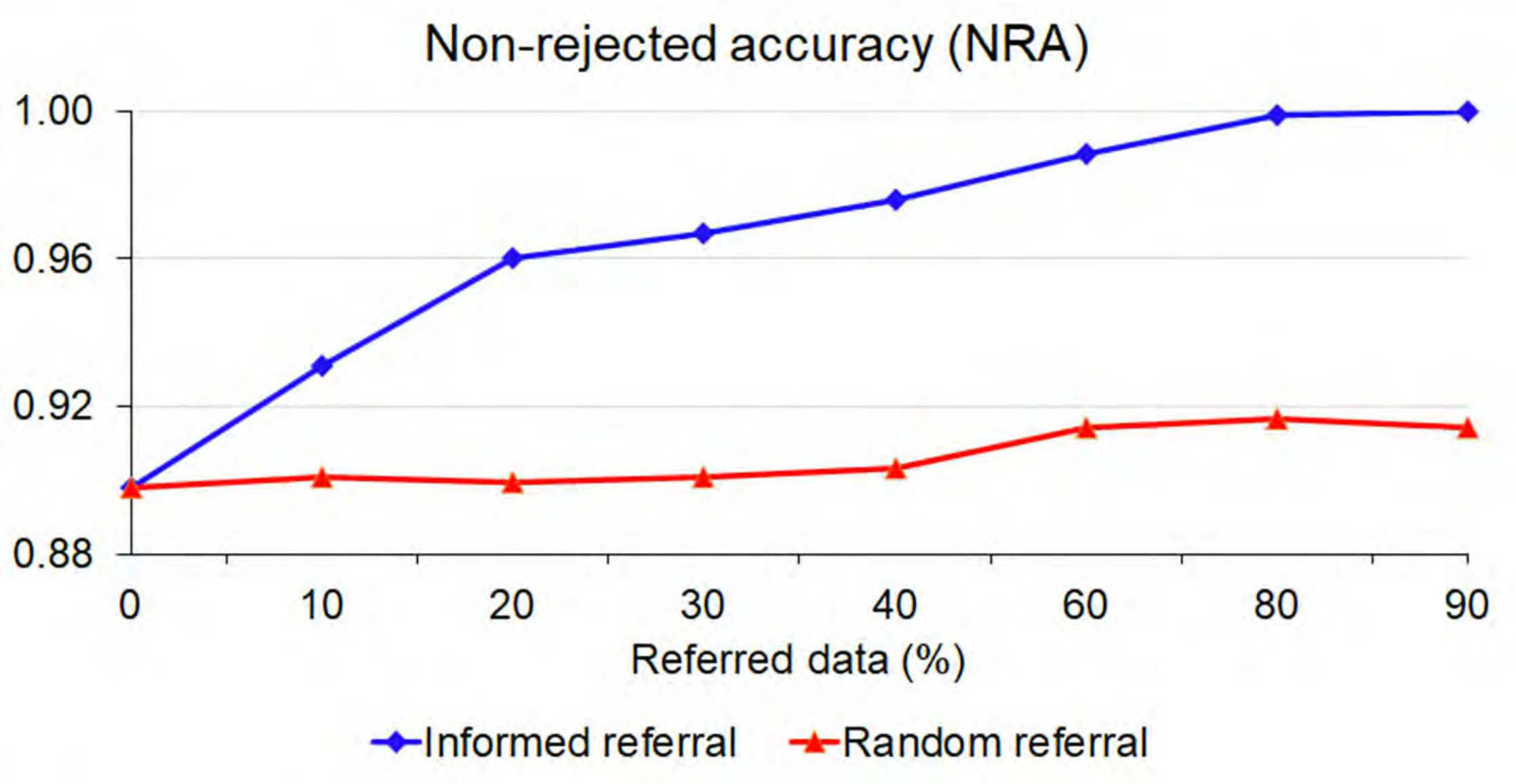}}
    \subfigure[]{\includegraphics[width=0.46\linewidth]{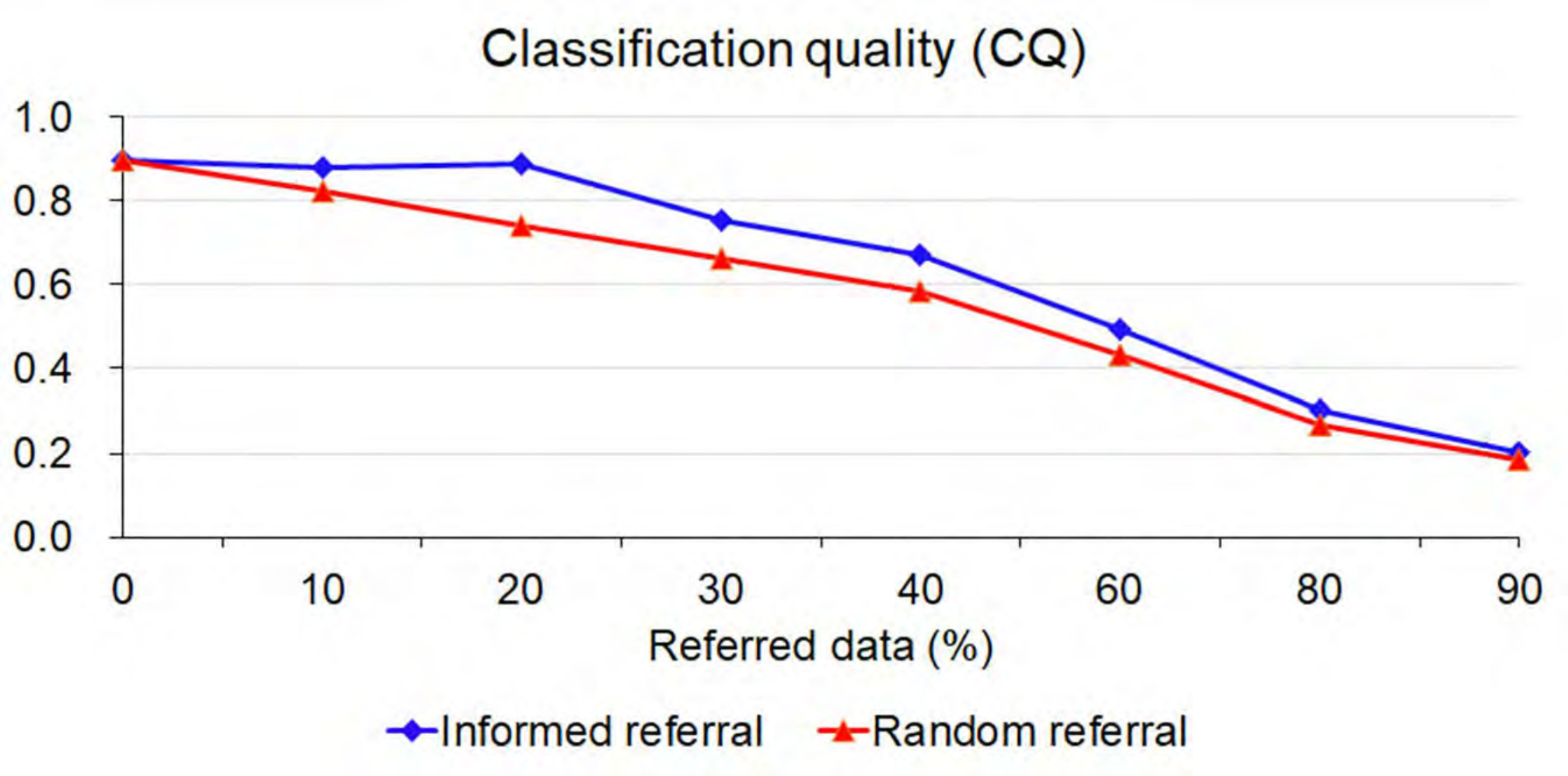}}
    \subfigure[]{\includegraphics[width=0.47\linewidth]{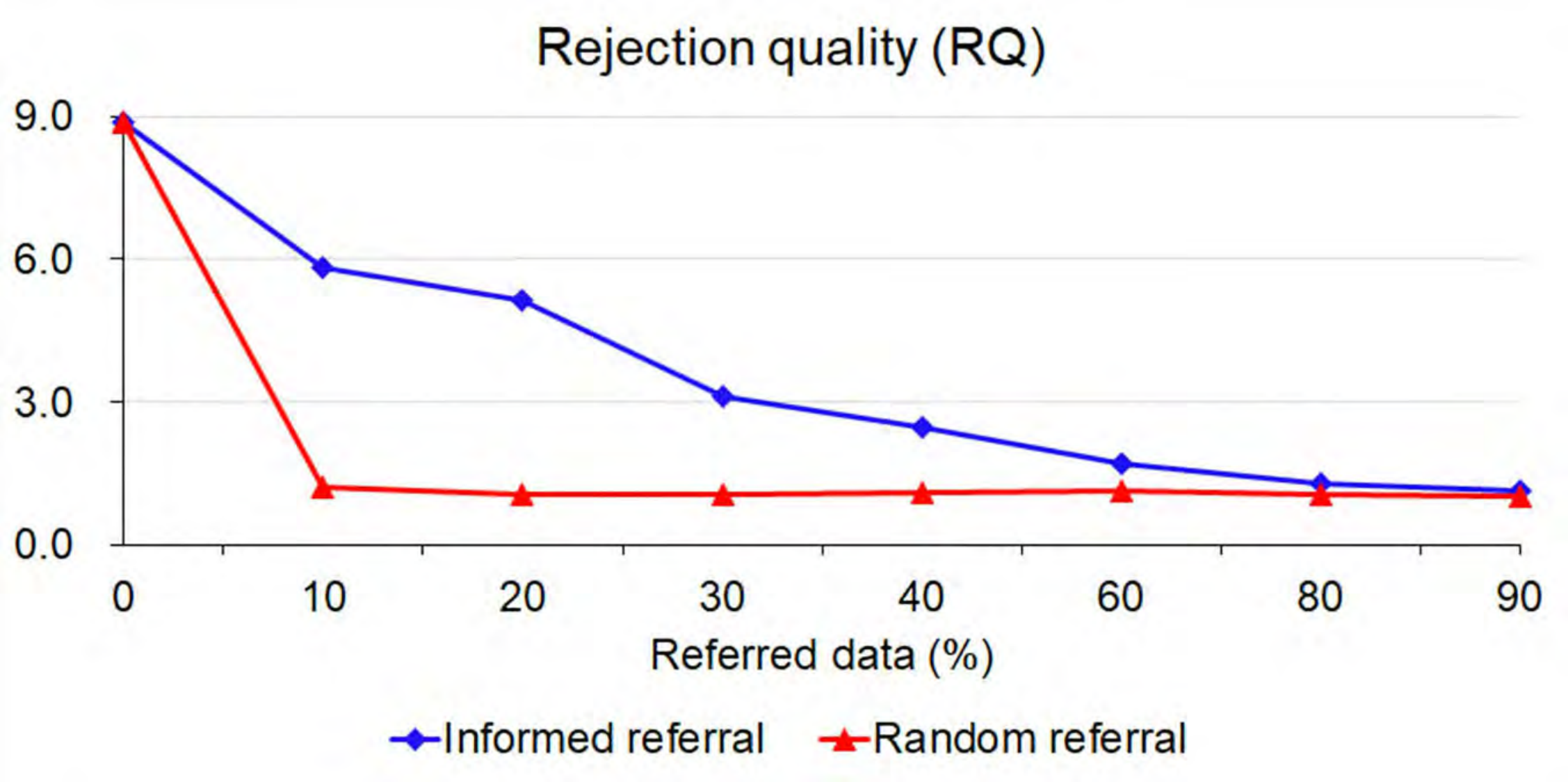}}
    \subfigure[]{\includegraphics[width=0.45\linewidth]{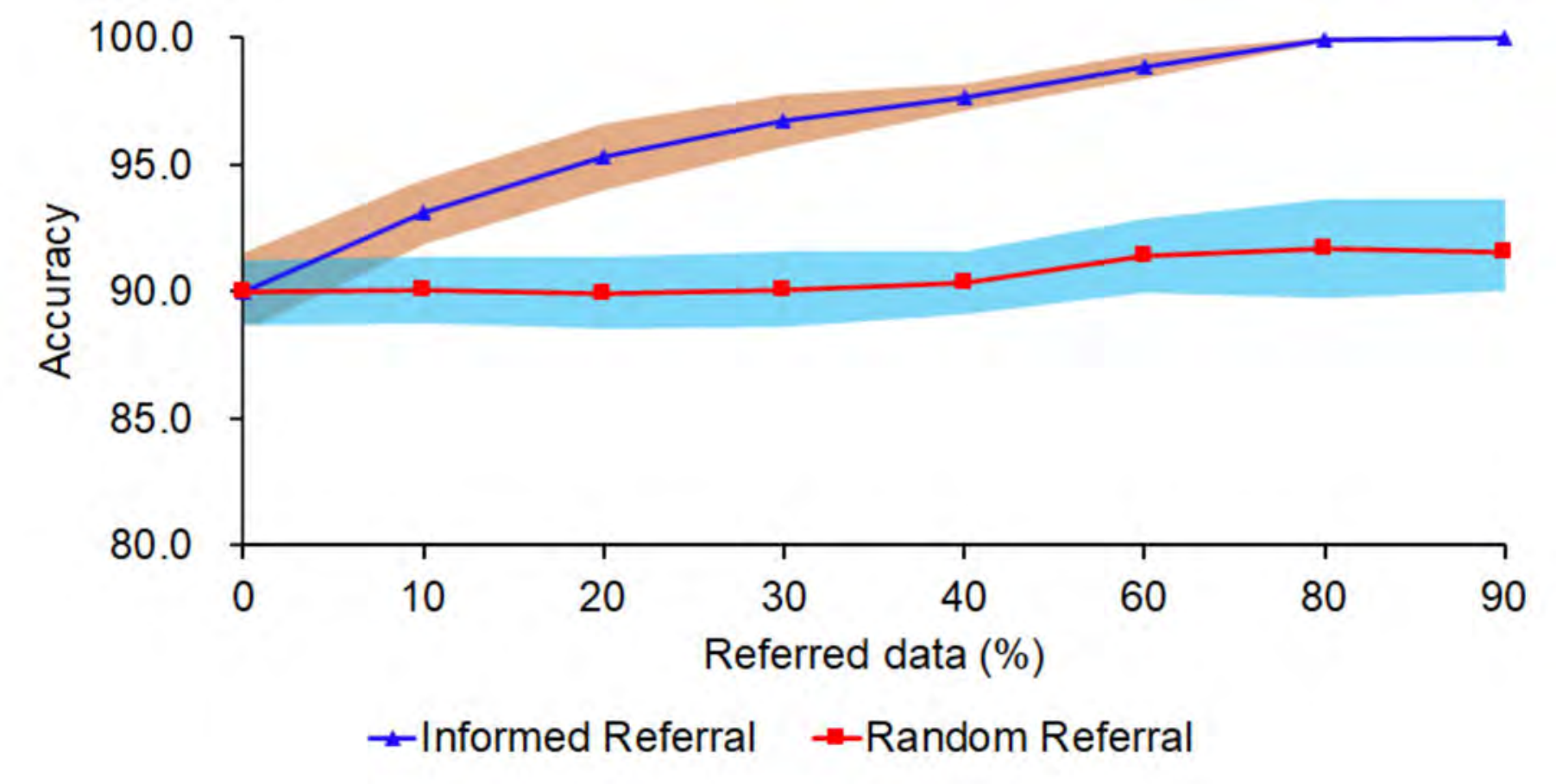}}
    \caption{Performance measure of informed vs random referral as a function of the rejected fraction. (a) Non-rejection accuracy ($NRA$), where the y-axis shows the accuracy of the non-rejected prediction. (b) classification quality ($CQ$), where the y-axis shows the number of points that are correctly/wrongly classified. (c) rejection quality ($RQ$), where the y-axis shows the number of points that are correctly/wrongly classified and not-rejected/rejected over the total samples. (d) the prediction accuracy over $N\%$ of retained data as a function of informed vs random referral. The standard deviation from the five-fold cross-validation is shown by the shaded region around the curves.}
    \label{fig:04}
\end{figure*}

We used two policies to assess the uncertainty in the classification with rejection. In the first policy, we chose an uncertainty threshold value from the range $\tau =\{0.08, 0.1, 0.2, 0.3\}$. The samples that do not meet the threshold are considered as uncertain and submitted to experts for further inspection. We compared the proposed architecture with~\cite{adhane2021deep,krizhevsky2012imagenet} as well as modified AlexNet. We modified the AlexNet~\cite{krizhevsky2012imagenet} in the same way as we did to~\cite{adhane2021deep} by adding the dropout layer after each convolution layer to the original AlexNet architecture. Table~\ref{tab:01} summarises the performance of various CNN architectures with varying uncertainty threshold values. In our experiments, we found that the minimum and maximum values of the estimated uncertainty (i.e., variance) in our dataset are 0.08 and 0.3, respectively. If we set the uncertainty threshold to a small value (e.g., 0.08 or 0.1), the classifier only accepted a few certain predictions and referred the rest to the expert, resulting in unfair yet great performance being reported. However, if we forced the model to accept a large uncertainty threshold (e.g. 0.3), the model accepted more incorrect predictions while only referring a few samples to the expert, resulting in lower classifier performance.

This is also evident in Table~\ref{tab:01}, where the threshold value and the number of rejected samples have an inverse relationship. To further study the impact of the step size on the performance of informed referral, we chose 0.1 as the second threshold, which differs from the previous and next values by 0.02 and 0.1, respectively. When the step size is set to 0.1, the number of rejected samples drops significantly. Indeed, the model can be adapted to the step size based on the application type or critical level. 

\begin{table*}[bp]
\centering
\caption{Comparison of informed and random referral for uncertain samples using class-wise recall, precision, and F1-score.}
\label{tab:02}
\begin{tabular}{lccccccccc}
\hline
 & \textbf{Number of samples} & \textbf{TP} & \textbf{TN} & \textbf{FP} & \textbf{FN} & \textbf{Precision} & \textbf{Recall} & \textbf{F1-score} & \textbf{Accuracy}\\ \hline
No referral & 1275 & 614 & 553 & 58 & 70 & 0.91 & 0.89 & 0.90 & 0.91\\
Random referral (20\%) & 1021 & 486 & 432 & 49 & 54 & 0.90 & 0.90 & 0.90 & 0.90\\
Informed referral (20\%) & 1021 & 506 & 487 & 20 & 28 & 0.96 & 0.94 & 0.95 & 0.97\\ \hline
\end{tabular}
\end{table*}

In the second policy, first, we ranked the prediction of all test data sets based on the estimated uncertainty. Then, to evaluate the model, we referred $N\%$ of the uncertain samples (i.e., referred data) to an expert. Finally, we assessed the performance of the model on the remaining data. We compared rank-based sampling (i.e., informed referral) with random-based sampling (i.e., random referral). As shown in Fig.~\ref{fig:04}, the informed referral outperforms when the model is allowed to refer more data to an expert. We can also see in Fig.~\ref{fig:04} that referring 10-20\% of samples to the expert improves the model's performance by up to 4\%. We further investigated the informed and random referral using class-wise recall, precision, and F1-score at the rate of 20\% (see Table~\ref{tab:02}). As shown in Table~\ref{tab:02}, ranking samples based on estimated uncertainty can significantly improve model performance.

Figure~\ref{fig:04}(a) shows the relationship between the percentage of data referred to an expert and the non-rejected accuracy (NRA). For example, if we ask an expert to evaluate 20\% of the most uncertain samples in the informed referral, the model will be assessed by the remaining 80\% of the certain samples. Therefore, the variance $(\sigma)$ will be less than the threshold $(\tau)$, resulting in an accurate classification with an NRA close to 1. However, in the case of random referral, if we ask an expert to evaluate 20\% of the samples at random, the model is assessed by the remaining 80\% of the samples, which are not necessarily certain or uncertain. In this case, the threshold will not satisfy the variance (uncertainty metric), and the $NRA$ rate will be reduced due to an increase in misclassified and rejected samples. The classification quality ($CQ$) and rejection quality ($RQ$) in Fig.~\ref{fig:04} (b) and (c) are more concerned with the number of accepted correctly classified samples and incorrectly classified samples referred to, respectively. Therefore, referring 80\% of total samples to an expert implies that the expert will inspect a large number of correctly classified samples, compromising the classifier's $CQ$ and $RQ$.

\begin{figure*}[!htbp]
    \xdef\xfigwd{\columnwidth}
    \centering
    \includegraphics[width=0.9\linewidth]{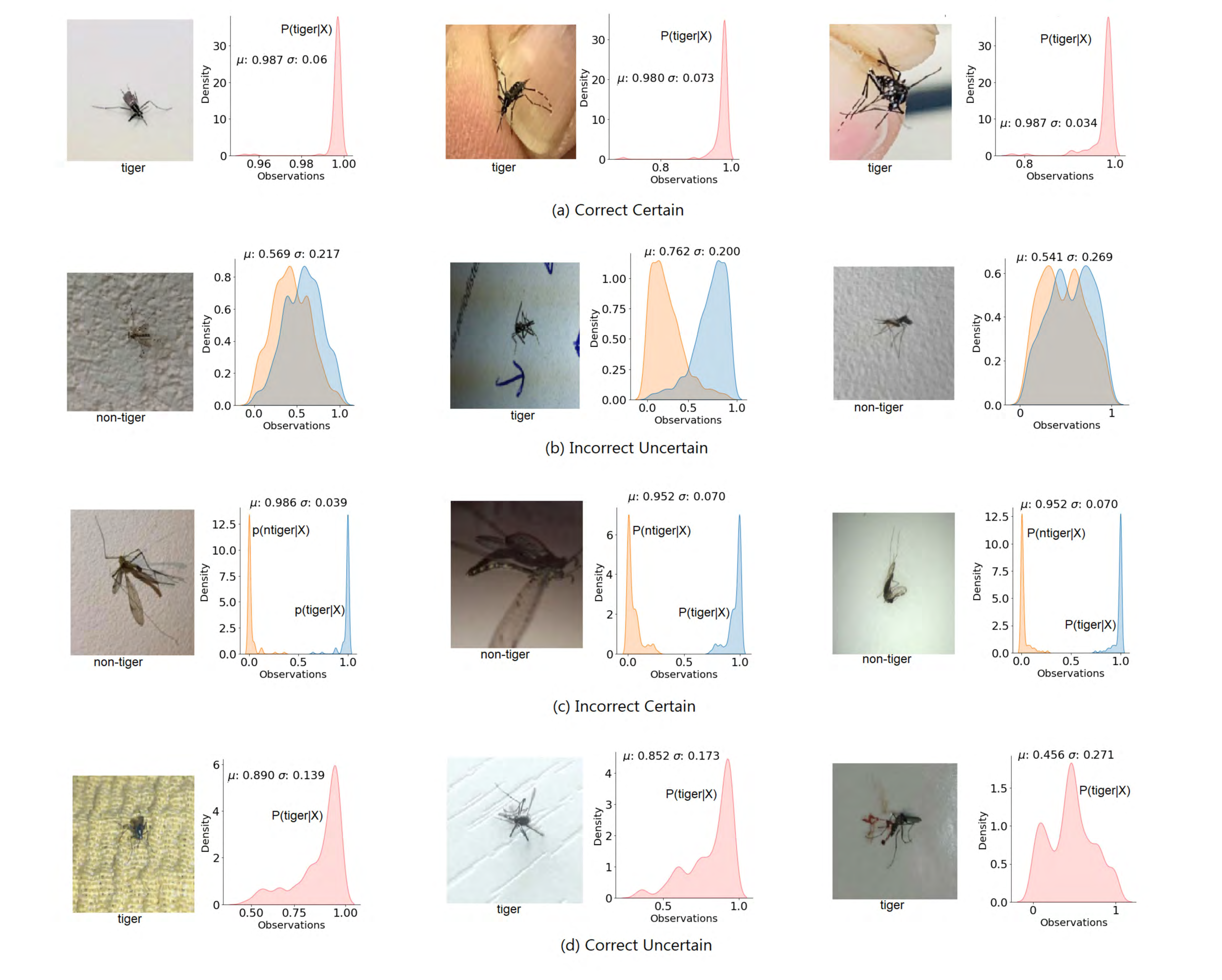} 
    \caption{Visualisation of the posterior distributions and associated uncertainty estimates for 12 samples at $\sigma \leq 0.1$. The distribution of correct predictions is shown in red, while the distribution of incorrectly classified samples is shown in blue.}
    \label{fig:05}
\end{figure*}

Depending on the uncertainty threshold, the outcome of predictive posterior distribution of our experiment can have one of the four states: \textit{correct certain}, \textit{correct uncertain}, \textit{incorrect uncertain}, and \textit{incorrect certain}. We visualised the histogram of the predictive posterior distribution for 12 inputs to better understand the classifier's output (See Fig.~\ref{fig:05}). In cases where there is only one histogram with a narrow tail (i.e., \textit{correct certain}), the classifier is certain in its prediction and accepts the decision. However, when the histogram has a broader tail (i.e., \textit{correct uncertain}), the classifier is uncertain in its prediction and refers the decision to the expert. Also, when there are two overlapped histograms (i.e., \textit{incorrect uncertain}), the classifier is uncertain in its prediction and refers the decision to the expert. When there are two non-overlapping histograms with narrow tails (i.e., \textit{incorrect certain}), the classifier is certain in its prediction and accepts the decision with a wrong prediction. Since the model is fed with images acquired in-the-wild, in several cases, it needs to process, for instance, (1) different objects in the image that are more salient than the mosquito and (2) mosquitoes with damaged body parts (see Fig.~\ref{fig:05} (b) and (d)). Such factors can contribute to uncertainty and variation in the results.

\subsubsection{Comparison with competing methods}
Table~\ref{tab:03} compares the proposed method with~\cite{adhane2021deep,pataki2021deep,12he2016deep} based on Accuracy and ROC AUC metrics. To validate the proposed method, we tested the model with 885 samples that met the $\sigma \leq 0.1$. Pataki et al.~\cite{pataki2021deep} evaluated their approach using a yearly-cross validation strategy in which they randomly selected 1000 samples from each year's data to train the model and tested it with the data of other years. We trained the fine-tuned version of VGG16~\cite{adhane2021deep} and ResNet50~\cite{12he2016deep} classifiers on verified images by experts using a five-fold cross-validation strategy.

\begin{table}[!htbp]
\centering
\caption{Performance of the proposed approach compared to competing methods considering Accuracy and ROC AUC.}
\label{tab:03}
\begin{tabular}{lccc}
\hline
\textbf{Architecture} & \textbf{Samples \#} & \textbf{Accuracy} & \textbf{ROC AUC} \\ \hline
Proposed method & 885 & 0.98 & 0.98 \\
Adhane et al.~\cite{adhane2021deep} & 1275 & 0.94 & 0.96 \\
ResNet50~\cite{12he2016deep} & 1275 & 0.93 & 0.95  \\
Pataki et al.~\cite{pataki2021deep} & 1000 & 0.88 & 0.91 \\ \hline
\end{tabular}
\end{table}

Table~\ref{tab:03} shows that the performance of the modified VGG16 is marginally higher than that of the standard VGG16~\cite{adhane2021deep} and ResNet50~\cite{12he2016deep,pataki2021deep}, even though it requires more trainable parameters than ResNet50. The main reason for this marginal superiority is the larger number of learning parameters, which allows the model to be fine-tuned to produce a more discriminative feature space when used in transfer learning.

\begin{table*}[bp]
\centering
\caption{Test accuracy as a function of the number of images acquired during each active learning iteration.}
\label{tab:04}
\begin{tabular}{rccccccccc}
\hline
\textbf{Training iteration}                 & \textbf{0} & \textbf{5}  & \textbf{10} & \textbf{15} & \textbf{20} & \textbf{25} & \textbf{30} & \textbf{35} & \textbf{40}  \\ \hline
\textbf{Labelled samples(\%)} & \textbf{6} & \textbf{17} & \textbf{29} & \textbf{41} & \textbf{52} & \textbf{64} & \textbf{75} & \textbf{87} & \textbf{100} \\ \hline
Proposed method                                         & 0.75       & 0.87        & 0.91        & 0.93        & 0.93        & 0.93        & 0.93        & 0.93        & 0.93         \\
Least Confidence                            & 0.75       & 0.87        & 0.89        & 0.92        & 0.92        & 0.93        & 0.93        & 0.93        & 0.93         \\
Random                                      & 0.77       & 0.85        & 0.87        & 0.88        & 0.89        & 0.92        & 0.92        & 0.92        & 0.93         \\ \hline
\end{tabular}
\end{table*}

\subsubsection{Explainability of classification with rejection}
\begin{figure*}[!htbp]
    \xdef\xfigwd{\columnwidth}
    \centering
    \subfigure[]{\includegraphics[width=0.45\linewidth]{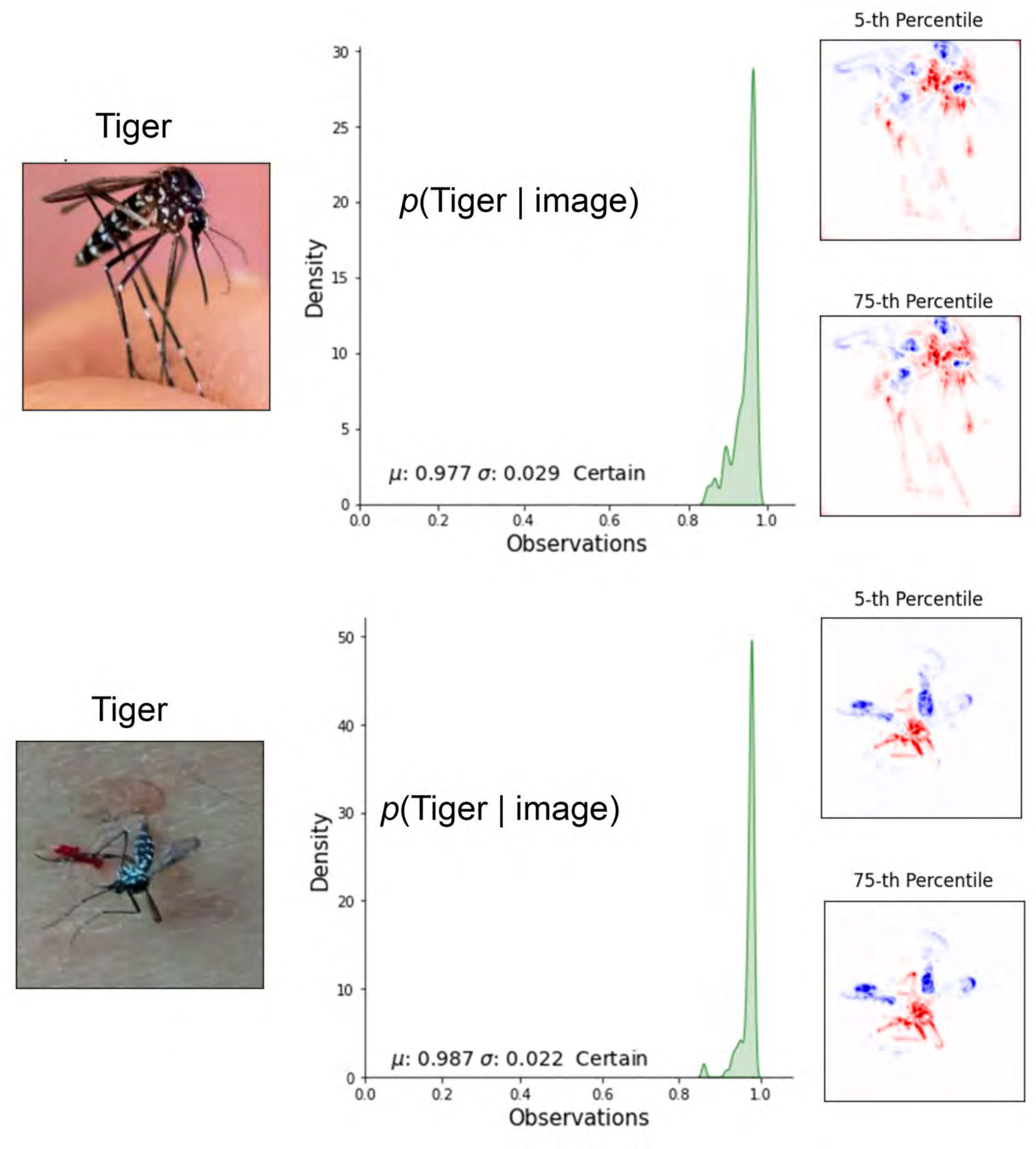}}
    \subfigure[]{\includegraphics[width=0.45\linewidth]{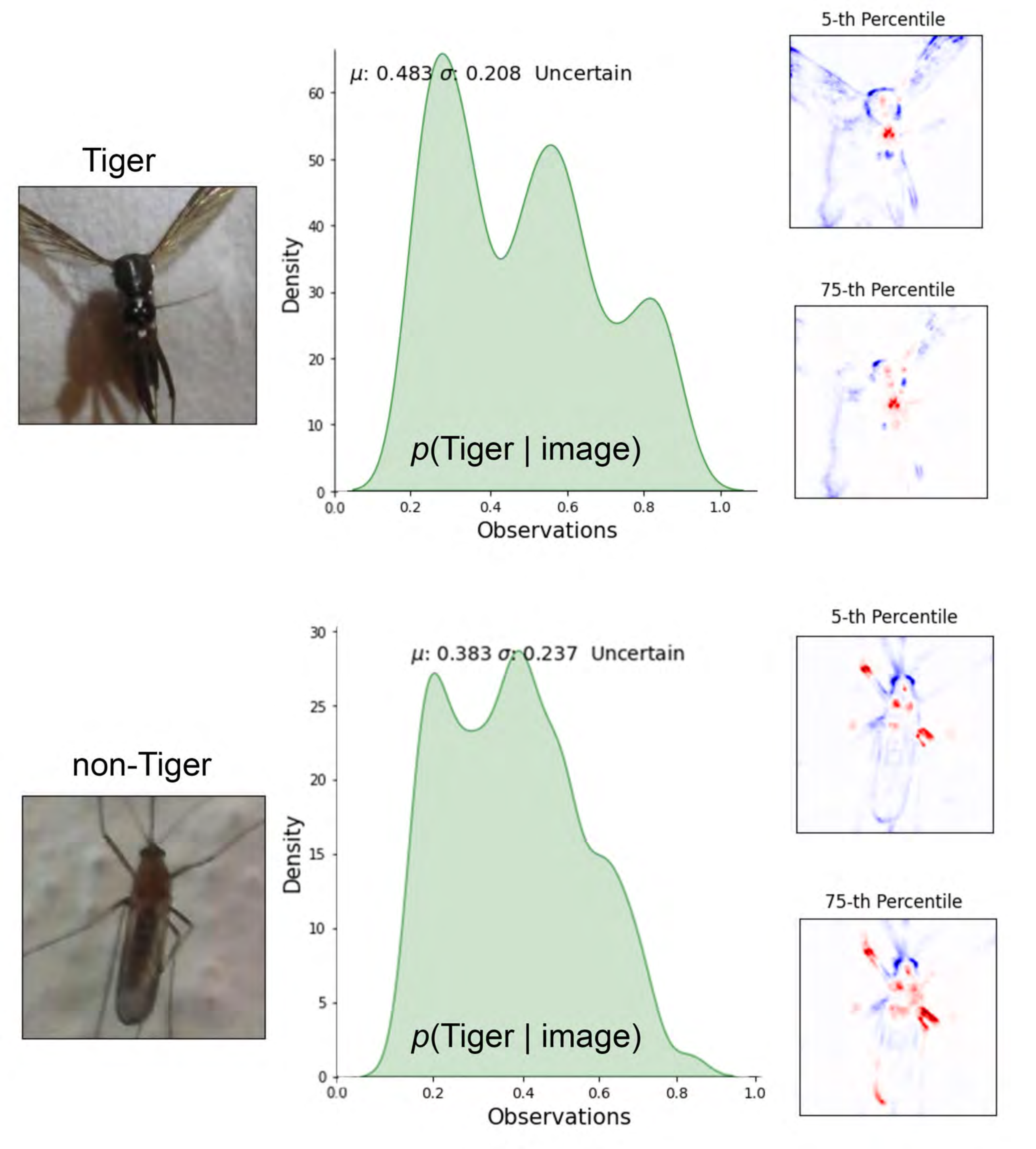}}
    \caption{Exemplary explanation of the prediction made by proposed method using B-LRP~\cite{45bykov2020much}. (a) Two samples of \textit{Aedes albopictus} (Tiger) mosquitoes where the model is certain in its prediction. The B-LRP highlights the thorax and legs as key discriminant features in the $5^{th}$ percentile and it highlights the regions that can be referred to as the second discriminant features in the $75^{th}$ percentile. (b) Two samples of \textit{Aedes albopictus} (Tiger) and non-\textit{Aedes albopictus} (non-Tiger) mosquitoes, where the model is uncertain in its prediction. Here, the B-LRP in the $5^{th}$ and $75^{th}$ is unable to highlight discriminative regions. Red pixels indicate positive relevance, and blue pixels represent negative relevance.}
    \label{fig:07}
\end{figure*}

In this part of experiment, we used B-LRP~\cite{45bykov2020much}, a Bayesian neural network explanation technique that generates saliency maps, to explain the uncertainties associated with the prediction and to identify certain and uncertain regions for the proposed classification with rejection. Fig.~\ref{fig:07} shows examples of pixel relevance visualisation using B-LRP for \textit{Aedes albopictus} and other types of mosquitoes. To visualise the exemplary explanations, we considered the $5^{th}$ and $75^{th}$ percentiles of pixel importance, where the former represents pixels for the most certain regions, and the latter illustrates pixels that are potentially relevant but have a degree of uncertainty.

The B-LRP in the $5^{th}$ percentile highlights the thorax and legs as key discriminant features, as also reported by~\cite{adhane2021deep}, while in the $75^{th}$ percentile, the regions that can be referred to as the second discriminant features of \textit{Aedes albopictus} were highlighted.

\subsubsection{Impact uncertainty in active learning framework}
\begin{figure}[!htbp]
    \centering
    \includegraphics[width=\linewidth]{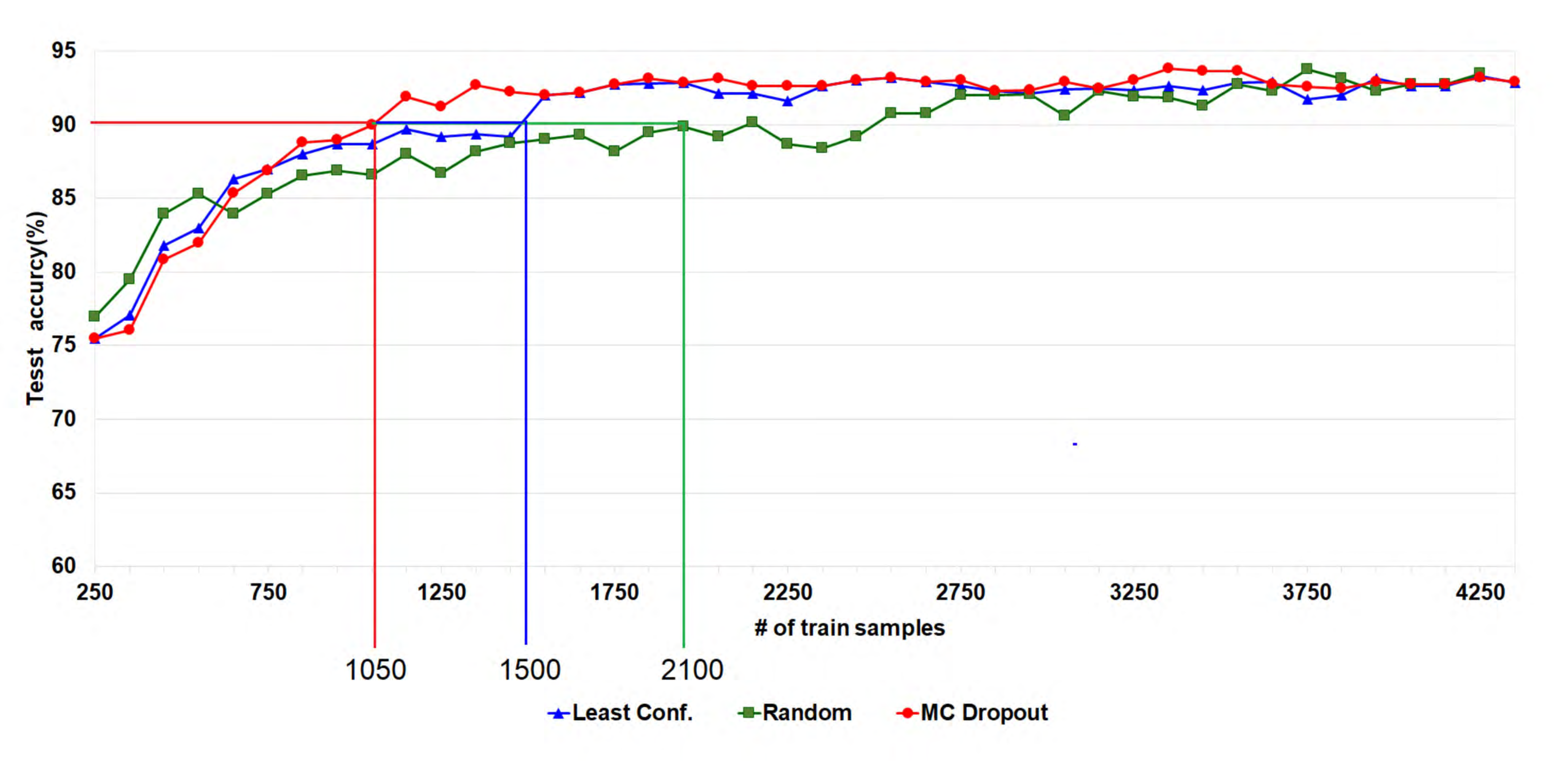}
    \caption{A comparison of three query strategies based on accuracy and the number of images acquired from the pool.}  
    \label{fig:08}
\end{figure}

We compared the Dropout-based query strategy to random and least-confidence query strategies in the automated classification of \textit{Aedes albopictus} images. According to the findings shown in Table~\ref{tab:04} and Fig.~\ref{fig:08}, the proposed framework reaches higher test accuracy with minimal labelling. Although the MC Dropout-based query strategy converges slower than other strategies at first, it gradually improves by starting with a limited number of labelled samples and continuing by labelling sufficiently training samples. A closer look at Fig.~\ref{fig:08} reveals that MC dropout-based query strategy can achieve a test accuracy of 90\% by using just 25\% of the available data, whereas the least-confidence and random query strategies require 34\% and 58\% of the available data, respectively. Thus, depending on the level of performance desired, uncertainty-based active learning has the ability to reduce data labelling costs.

\section{Conclusion}\label{sec:conclusion}
The re-emergence of mosquito-borne diseases (MBDs) kills hundreds of thousands of people each year by carrying and transmitting infectious and contagious diseases to humans through direct or indirect contact. Several studies have used convolutional neural networks (CNNs) to recognise mosquitoes in images provided by projects such as Mosquito Alert to help entomologists in identifying, monitoring, and managing MBD. However, not only are these methods associated with incorrect predictions, but they also require a large amount of manually annotated data.

In this study, we have proposed two significant contributions to minimise human intervention while maximising model accuracy in the Mosquito Alert project, which has a large amount of contributed data. First, we have included the Monte Carlo (MC) dropout to incorporate uncertainty in the prediction. We have kept dropout active during testing and made $T$ stochastic forward passes through the network. Averaging the softmax vector obtained from $T$ forward passes provides the final posterior probability distribution for a given input, with the variance of this vector serving as model uncertainty. In this way, we could identify samples that may have been misclassified. We have then ranked the predicted samples and sent only samples with the highest level of uncertainty to an expert. Second, we have proposed an active learning scheme that initiates with a subset of manually annotated samples and gradually adds more unlabelled data. The uncertainty estimation guides this process and sends only the labels that are more likely to be inaccurate to the expert. 

Incorporating the uncertainty measure has a threefold advantage in the automated classification of \textit{Aedes albopictus}: (1) reducing the need for annotation of the entire database by assisting annotators in an active learning framework during training; (2) sorting the most uncertain samples at the time of inference to request re-validation by entomologists in order to reduce the number of mislabeled/misclassified samples; and (3) improving the performance of our previously modified VGG16~\cite{adhane2021deep} by 4\% using Monte Carlo Dropout uncertainty estimation. Furthermore, we explore the causes of model uncertainty by using a Bayesian visual explanation method to highlight regions in certain and uncertain predictions. In the most certain predictions, the model highlights the abdomen, thorax, and legs of the \textit{Aedes albopictus}, whereas, in the prediction associated with high variance, it highlights the non-discriminant regions.


\section*{Acknowledgment}
This research was supported by ``RTI2018-095232-B-C22" grant from the Spanish Ministry of Science, Innovation and Universities (FEDER funds), and NVIDIA Hardware grant program. 

We also acknowledge the Mosquito team (http://www.mosquitoalert.com/en/about-us/team/) for their work in keeping the system operative, even in harsh financial times, and most especially the team of volunteer entomology experts that have validated mosquito pictures from Mosquito Alert during the period 2014 to 2019: Mikel Bengoa, Sara Delacour, Ignacio Ruiz, Maria Àngeles Puig, Pedro María Alarcon-Elbal, Rosario Melero-Alcíbal, Simone Mariani, and Santi Escartin. Finally, we would like to thank the Mosquito Alert community (anonymous citizens) who have participated year by year, making all this data collection system worth it.

\bibliographystyle{IEEEtran.bst}
\bibliography{reference.bib}

\begin{IEEEbiography}[{\includegraphics[width=1in,height=1.25in,clip,keepaspectratio]{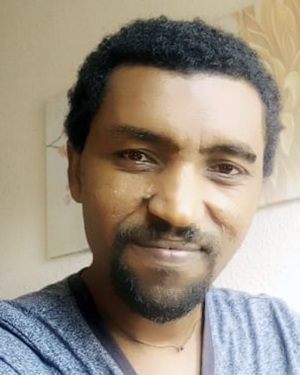}}]{Gereziher Adhane} is currently a PhD student at Universitat Oberta de Catalunya, Spain. He obtained his MSc from Osmania University (India) in 2013/14. His research interests includes deep learning, computer vision and fairness in AI.
\end{IEEEbiography}

\begin{IEEEbiography}[{\includegraphics[width=1in,height=1.25in,clip,keepaspectratio]{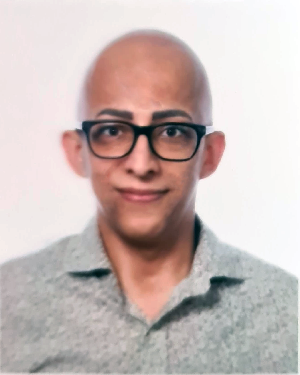}}]{Mohammad Mahdi Dehshibi}
is currently a postdoctoral research fellow at Universitat Oberta de Catalunya, Spain. He obtained the PhD from IAU (Iran) in 2017. He was also a visiting researcher at Unconventional Computing Lab, UWE, Bristol, U.K. He has contributed to over 60 papers published in scientific Journals and International Conferences. His research interests include Affective and Unconventional Computing. 
\end{IEEEbiography}

\begin{IEEEbiography}[{\includegraphics[width=1in,height=1.25in,clip,keepaspectratio]{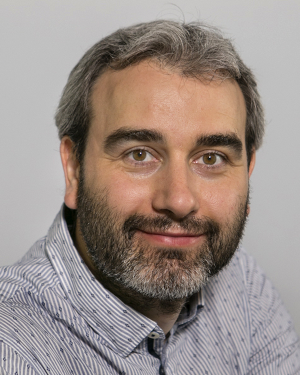}}]{David Masip} is Professor in the Computer Science Multimedia and Telecommunications  Department, Universitat Oberta de Catalunya since February 2007 and Director of the Doctoral School since 2015. He is the director of the Scene Understanding and Artificial Intelligence Lab and member of the BCN Perceptual Computing Lab. He studied Computer Vision at the Universitat Autonoma de Barcelona. He received his PhD in 2005 and was awarded for the best thesis in the Computer Science.
\end{IEEEbiography}
\vfill
\end{document}